\documentclass[preprint]{article}

\usepackage{neurips_2024}

\usepackage[utf8]{inputenc} 
\usepackage[T1]{fontenc}    
\usepackage{hyperref}       
\usepackage{url}            
\usepackage{booktabs}       
\usepackage{amsfonts}       
\usepackage{amsmath}
\usepackage{graphicx}
\usepackage{wrapfig}
\usepackage{multirow}
\usepackage{float}
\usepackage{nicefrac}       
\usepackage{microtype}      
\usepackage{xcolor}         
 \usepackage{natbib}
\setcitestyle{numbers,square}

\title{EEPNet: Efficient Edge Pixel-based Matching
Network for Cross-Modal Dynamic Registration
between LiDAR and Camera}

\author{
  Yuanchao Yue$^1$, Hui Yuan$^1$\thanks{Corresponding author. \texttt{huiyuan@sdu.edu.cn}},  Suai Li$^1$,  Qi Jiang$^1$\\
 $^1$School of Control Science and Engineering, Shandong University  \\
\texttt{202234945@email.com}, \texttt{huiyuan@sdu.edu.cn}\\\texttt{shuaili@sdu.edu.cn}, \texttt{jiangqi@sdu.edu.cn}
}

\begin{document}

\maketitle

\begin{abstract}
Multisensor fusion is essential for autonomous vehicles to accurately perceive, analyze, and plan their trajectories within complex environments. This typically involves the integration of data from LiDAR sensors and cameras, which necessitates high-precision and real-time registration. Current methods for registering LiDAR point clouds with images face significant challenges due to inherent modality differences and computational overhead. To address these issues, we propose EEPNet, an advanced network that leverages reflectance maps obtained from point cloud projections to enhance registration accuracy. The introduction of point cloud projections substantially mitigates cross-modality differences at the network input level, while the inclusion of reflectance data improves performance in scenarios with limited spatial information of point cloud within the camera's field of view. Furthermore, by employing edge pixels for feature matching and incorporating an efficient matching optimization layer, EEPNet markedly accelerates real-time registration tasks. Experimental validation demonstrates that EEPNet achieves superior accuracy and efficiency compared to state-of-the-art methods. Our contributions offer significant advancements in autonomous perception systems, paving the way for robust and efficient sensor fusion in real-world applications.
\end{abstract}

\section{Introduction}

In recent years, advances in autonomous driving \cite{Grigorescu_2019}\cite{5970711} and robotics \cite{8633982} have spurred significant developments in sensor technology and data fusion techniques. Multisensor fusion has emerged as a crucial technology for achieving autonomous driving capabilities. It enables vehicles to perceive, understand, and plan their trajectories within their surrounding environment. Autonomous vehicles rely on sophisticated perception systems to acquire accurate and comprehensive environmental information, underscoring the critical importance of developing advanced perception systems to progress autonomous driving technology.\\
In real-world autonomous driving, sensor fusion entails integrating data from LiDAR sensors, such as the Velodyne HDL-64E, and cameras. This fusion necessitates a precise rigid transformation matrix between the LiDAR point cloud data and the camera image data, enabling the alignment of data from different sensors into a unified reference frame. However, practical applications encounter challenges such as relative movement between the LiDAR and the camera induced by human factors or mechanical vibrations, leading to positional deviations. Additionally, dynamic changes in sensor positions during real-time operation are commonplace. In such cases, real-time registration is crucial in adjusting for these positional discrepancies, ensuring accurate sensor fusion in autonomous driving systems.\\
Common approaches to addressing the registration problem typically involve establishing correspondences between three-dimensional (3D) points and two-dimensional (2D) pixels, followed by employing techniques such as Efficient Perspective-n-Point (EPnP) \cite{Lepetit2009} to estimate the corresponding pose matrix. Traditional sensor calibration methods, for instance, rely on 2D-3D correspondences derived from point and line features on calibration boards to align camera images with LiDAR point clouds. However, these methods are limited by the requirement for fixed targets and are unsuitable for real-time registration.\\
Continuous advancements have been observed in deep learning methods for registering images with point clouds.  These approaches typically employ neural network-based feature extractors to capture high-dimensional features from 2D images and 3D point clouds. Subsequently, similarity measures in this high-dimensional feature space are utilized to establish 2D-3D correspondences, facilitating registration. However, the inherent modality differences between the 3D spatial data acquired by LiDAR sensors and the 2D images captured by cameras present a big challenge. Some existing cross-modality registration methods \cite{li2021deepi2p}\cite{Ren_2023}\cite{NEURIPS2023_a0a53fef} introduce specialized network structures to mitigate cross-dimensional disparities, albeit at the cost of increased computational overhead and reduced processing speed.\\
Current methods primarily rely on matching spatial features extracted from point clouds with color features extracted from images. However, this approach may suffer from significant performance degradation when the spatial information of the point cloud corresponding to the camera's field of view is not sufficiently rich. Despite the potential to enhance relevant features by incorporating reflectance information during point cloud processing, thereby improving network performance in such scenarios, the inherent limitations of existing point cloud feature extractors may prevent this approach from being fully effective.\\
In response to these challenges, we present the EEPNet which substitutes the original point cloud with a reflectance map obtained via specific projection techniques for matching, thereby mitigating cross-dimensional disparities while enhancing network computational efficiency. Moreover, integrating reflectance information improves registration success rates in scenarios where the spatial information of the point cloud within the camera's field of view is not sufficiently rich. EEPNet uses edge pixels from the reflectance map and the camera image for feature matching, transforming the 2D pixel- 3D point matching problem into a 2D pixel- 2D pixel matching problem. In addition, the inclusion of a matching optimization layer during the matching stage significantly enhances registration accuracy and computational efficiency. Experimental validation demonstrates superior accuracy and computational efficiency of EEPNet, compared to state-of-the-art methods, making it well-suited for real-time registration. Our contributions are summarized as follows:
\begin{itemize}
    \item We propose an algorithm for real-time registration between LiDAR point clouds and camera images, leveraging point cloud projections. This algorithm explores the feasibility of utilizing projection for registration purposes.
    \item We propose a method to project point clouds into 2D reflectance maps, incorporating reflectance information into the registration task. This approach addresses performance issues when point cloud spatial information within the camera's field of view is insufficient.  
    \item We propose a method leveraging features of edge pixels for matching and integrating a matching optimization network to substantially enhance computational efficiency while maintaining feature saliency.
\end{itemize}
\section{Related Work}
\subsection{Same-Modality Registration}
\textbf{Image Registration.} 
Traditional image matching typically involves feature point matching, where 2D feature point correspondences are obtained. Subsequently, transformations between different images are derived using methods such as PnP \cite{Lepetit2009} or Bundle Adjustment \cite{10.1007/3-540-44480-7_21}. Commonly used feature descriptors include SIFT \cite{790410}, SURF \cite{BAY2008346}, and ORB \cite{6126544}. More recently, SuperGlue \cite{sarlin2020superglue} has been introduced, which leverages attention mechanisms and the Sinkhorn algorithm \cite{Sinkhorn1967ConcerningNM} to solve the optimal transport problem \cite{MAL-073}, achieving outstanding performance in matching tasks. LightGlue \cite{lindenberger2023lightglue} optimizes the computational complexity and efficiency of each module in SuperGlue, introducing a simpler matching optimization layer. GlueStick\cite{pautrat2023gluestick} is an enhancement over SuperGlue that incorporates line features to improve matching performance. \\
\textbf{Point Cloud Registration.}
Point cloud registration can be classified into two main categories: fine registration and coarse registration. Fine registration is often accomplished using the Iterative Closest Point (ICP) algorithm \cite{121791}. However, ICP is susceptible to local minima due to the uncertainty of the initial state. To mitigate this issue, various variants of the ICP algorithm \cite{6751291}\cite{pavlov2017aaicp} and the Normal Distributions Transform (NDT) algorithm \cite{1249285} were developed. Coarse registration methods also rely on correspondence matching. For example, D3Feat \cite{9157624} utilizes a fully convolutional network based on KPConv \cite{9010002} to describe the features of 3D local key points. Zhang \textit{et al.}\cite{zhang2023lidargenerated} proposed a method that utilizes images generated by LiDAR to accomplish the registration task of 3D point clouds.\\
\subsection{Cross-Modality Registration}
\textbf{Extrinsic Calibration Methods.} Current calibration methods focus on the task of self-calibration between LiDAR and the camera. These methods typically operate under the assumption that the positions of the sensors remain fixed during system operation, thus performing registration only once for each sequence of data. Unlike traditional methods that rely on calibration boards for sensor calibration \cite{8593660}\cite{10099031}, recent approaches leverage deep learning techniques for sensor data registration. For instance, techniques like LCCNet \cite{lv2021lccnet} correct extrinsic parameters by projecting point clouds onto the camera plane using an imprecise initial extrinsic matrix \cite{7995968}\cite{8593693}\cite{9206138}. DEdgeNet \cite{10160910} improves upon LCCNet by incorporating depth-discontinuous edges to enhance registration accuracy. ATOP\cite{9802778} introduces an object-level matching network to accomplish calibration tasks. Additionally, RGKCNet \cite{9623545}, proposed by Ye \textit{et al.}, utilizes key points from images and point clouds for online calibration. However, these methods often assume that the initial extrinsic parameters deviate only slightly from the actual values. They may fail in scenarios where there is a significant discrepancy between the initial and actual extrinsic parameters, or when no initial parameters are provided.\\
\textbf{Image-to-Point Cloud Registration.} Recent advancements in 2D-3D cross-domain descriptor methods \cite{8794415}\cite{pham2019lcd}\cite{Wang_2021_ICCV} have facilitated the utilization of EPnP for point cloud and image registration. For instance, approaches like Go-Match \cite{zhou2022geometry} have demonstrated that attention mechanisms can effectively alleviate discrepancies between cross-dimensional features. Methods such as DeepI2P \cite{li2021deepi2p}, CorrI2P \cite{Ren_2023}, and VP2P-Match \cite{NEURIPS2023_a0a53fef} have successfully employed EPnP to tackle challenges arising from significant extrinsic parameter deviations. Wang \textit{et al.}\cite{wang2023endtoend} proposed I2PNet, which designed an end-to-end coarse-to-fine 2D-3D registration architecture to further improve registration accuracy. These methods mitigate cross-modal discrepancies by integrating networks that enhance feature information or by introducing branches that enhance feature similarity. This enhances the saliency of cross-modal features during the matching process, resulting in superior matching performance. However, the inclusion of additional network structures may degrade computational efficiency, rendering real-time and high-precision registration unattainable.\\
\section{Method}
\textbf{Problem statement.} For a 2D pixel \((u_i, v_i)\) in an image and a 3D point \((x_i, y_i, z_i)\) in a point cloud, we can use a rigid transformation matrix \(\mathbf{T}\) to  project the 3D point to the 2D pixel. The matrix \(\mathbf{T}\) can be decomposed into a translation vector \(\mathbf{t}\in\mathbb{R}^3\) and a rotation matrix \(\mathbf{R}\in{SO}(3)\):

\[
\lambda \begin{bmatrix} u_i \\ v_i \\ 1 \end{bmatrix} = \mathbf{K} \mathbf{T} \begin{bmatrix} x_i \\ y_i \\ z_i \\ 1 \end{bmatrix}, \tag{1}
\]

where \(\mathbf{K}\) is the camera intrinsic matrix, and \(\lambda\) represents a scaling factor.\\
In this section, we begin by explaining how to convert a point cloud into a reflectance map. Then, we discuss how to use feature extraction networks to extract edge pixels and their features from the reflectance map and the camera image. After that, we introduce a matching optimization network to establish point correspondences. Finally, we use the EPnP algorithm to compute the transformation matrix for accurate registration.
\begin{figure}
    \centering
    \includegraphics[width=0.85\linewidth]{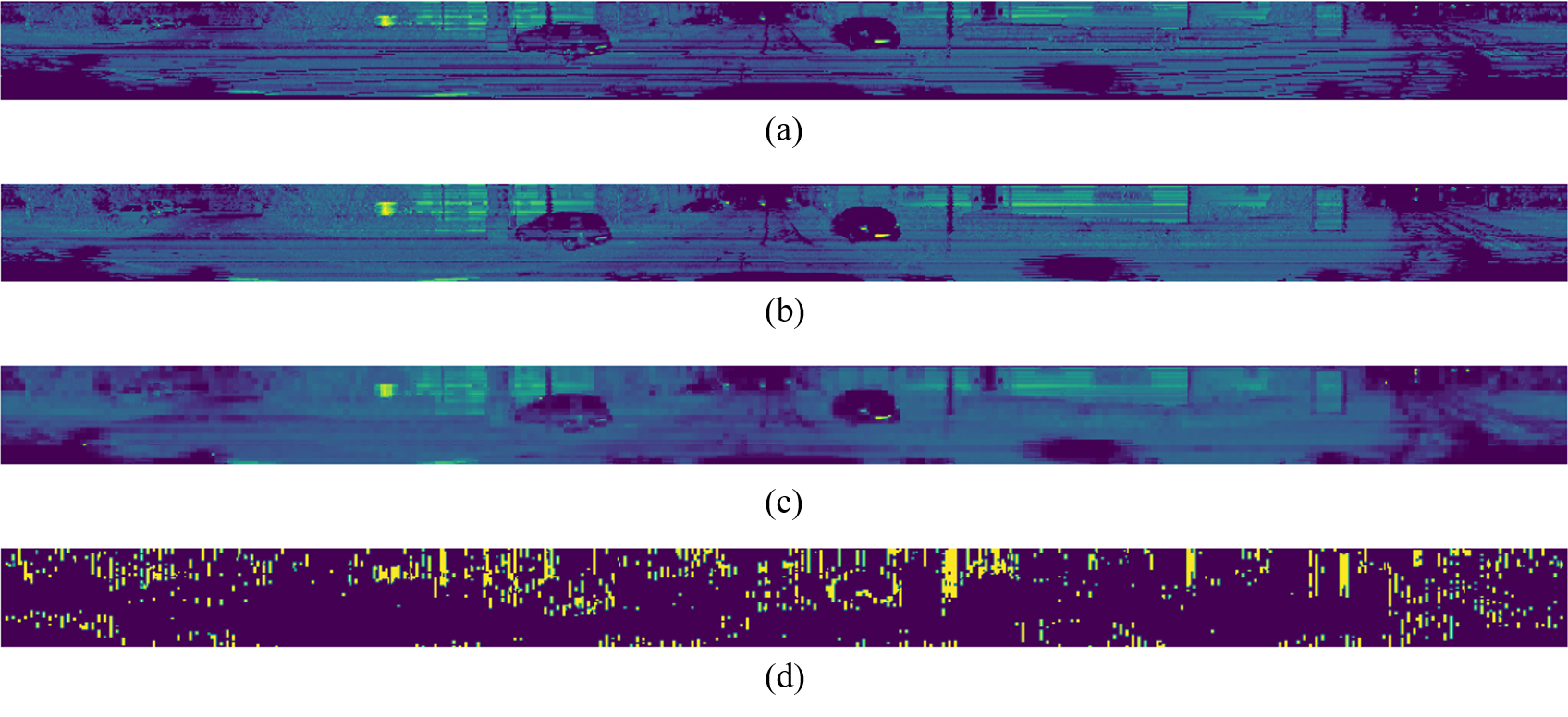}
    \caption{Visualization depicting the processing stages involved in projecting a 3D point cloud into a reflectance map and the following processing. (a) Projection map obtained through correlation with the spherical coordinate \(\theta\) and the vertical axis. (b) Projection map obtained through correlation with the LiDAR's LaserID on the vertical axis. (c) Post-processing of the projection map with wavelet filter, derived from (b). (d) Visualization of edge pixels extracted from (c) using the edge extraction algorithm.}   
    \label{fig:f1}
\end{figure}
\subsection{Reflectance Map}
When point clouds are collected by LiDAR, the reflectance information can be obtained by measuring the intensity of the reflected light from the target surface, while the depth information is determined by the distance from the LiDAR to the object. Therefore, the reflectance information is very similar to the optical information in camera images. We opted to project the reflectance information from the point cloud to obtain a reflectance map. Using reflectance information can provide sufficient optical data to achieve registration even when the spatial information of the point cloud within the camera's field of view is limited, thereby enhancing the system's robustness.
Common methods for projecting a 3D point cloud onto a 2D map typically include bird's-eye view(BEV) projection\cite{10321736}, cylindrical projection, and spherical projection. Among these methods, spherical projection closely resembles the image processing of a camera. \\
The spherical projection transforms the point cloud from Cartesian coordinates to spherical coordinates. The horizontal axis of the projected image corresponds to the angle \( \phi \) in the spherical coordinate system, calculated as \(\phi = \arctan{\left(\frac{y}{x}\right)}\), while the vertical axis corresponds to the \( \theta \) angle, calculated as \(\theta = \arccos{\left(\frac{z}{r}\right)}\), where the radius \( r \) is calculated using the formula \( r = \sqrt{x^2 + y^2 + z^2} \).
For common projection methods related to the vertical axis and angle \(\phi\), as well as the horizontal axis and angle \(\theta\), as illustrated in Fig. \ref{fig:f1}, panel (a). First, due to variations in the \(\theta\) angles between the laser emitters of different LiDAR models, the projected images may contain void regions. This issue was also investigated and visually demonstrated by Ren \textit{et al.}\cite{electronics10111224} across different LiDAR models. To address this challenge and facilitate future mappings between 3D point clouds and 2D projected maps, our approach leverages the LaserID of each laser emitter from the LiDAR as the vertical axis and the \(\phi\) angle as the horizontal axis for projection. The resulting projected map is depicted in Fig. \ref{fig:f1}, panel (b). It should be noted that during the projection process, we record the mapping relationship between the points in the point cloud and the pixels in the reflectance map, enabling us to identify the corresponding points in the 3D point cloud through the pixels in the reflectance map. This relationship is recorded as a projection matrix.
\subsection{Feature Extraction Network}
\subsubsection{Edge Pixel Extraction}
Drawing inspiration from the SuperPoint\cite{detone2018superpoint}, we incorporate high-dimensional features extracted from manually selected edge pixels into the proposed feature extraction network. To accomplish this, we utilize the method proposed by Gao \textit{et al.}\cite{5563693}, employing an improved Sobel edge detection technique to extract edge pixels from both the reflectance map and the camera image. Wavelet filter is used to mitigate the effects of high-frequency noise that exists in the reflectance maps generated during the LiDAR acquisition process. These noise artifacts, stemming from various sources of error during data collection, can compromise the accuracy of edge detection algorithms by erroneously identifying noise points as edges. Before edge extraction, a wavelet filter is applied to the reflectance maps to suppress high-frequency noise components. Specifically, a wavelet transform is performed on the image, followed by thresholding to remove high-frequency components below a predetermined threshold. This filter process smooths the image while preserving its essential structures and features. The filtered image, exemplified in (c) of Fig. \ref{fig:f1}, exhibits reduced noise and improved suitability for subsequent edge detection tasks.
We utilize Sobel operators specialized in vertical edge detection, which are adapted to accommodate the lateral scanning characteristics of LiDAR. Specifically, we define the following operators:

\[
\text{Sobel}_{left} = \left[ \begin{array}{ccc}
1 & 0 & -1 \\
2 & 0 & -2 \\
1 & 0 & -1
\end{array} \right], \text{Sobel}_{right} = \left[ \begin{array}{ccc}
-1 & 0 & 1 \\
-2 & 0 & 2 \\
-1 & 0 & 1
\end{array} \right].\tag{3}
\]

\begin{figure}[H]
    \centering
    \includegraphics[width=0.75\linewidth]{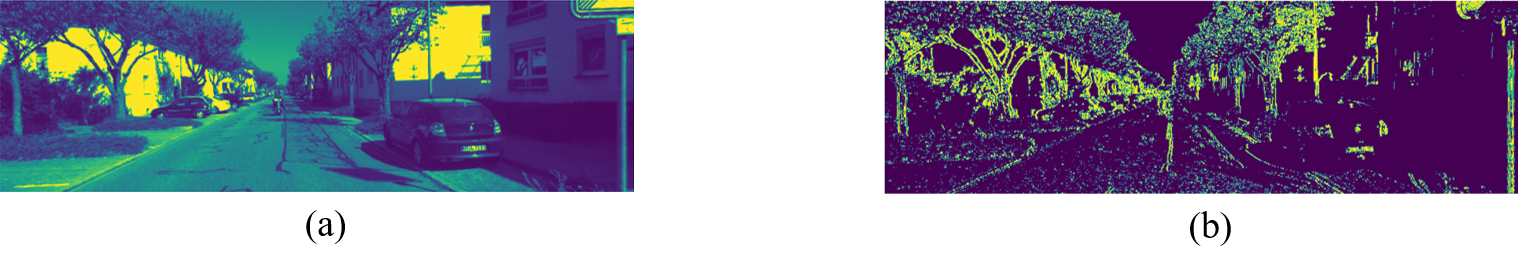}
    \caption{Visualization of image processing on the camera image. (a) Single-channel grayscale image of the camera image's R channel. (b) Visualization of edge pixels obtained after edge detection in (a). }
    \label{fig:f2}
\end{figure}

These Sobel operators are designed to capture edge features by the horizontal scanning pattern of the LiDAR sensor. \(\text{Sobel}_{left}\) detects vertical edges on the left side, while \(\text{Sobel}_{right}\) detects vertical edges on the right side. The use of these operators facilitates accurate edge detection in LiDAR data, enabling effective analysis of the surrounding environment. 
For the reflectance map, we employ an edge detection algorithm with a wavelet filter to extract a set of edge pixels containing \(N_r\) pixels, denoted as \(\mathbf{O}_r \in \mathbb{R}^{N_r \times 2}\). The visualization of edge pixels in the reflectance map is depicted in Fig. \ref{fig:f1}, panel (d). For the camera image, we perform edge detection, resulting in \(N_c\) pixels. We obtain the set of edge pixels in the camera image as \(\mathbf{O}_c \in \mathbb{R}^{N_c \times 2}\), with the visualization of edge pixels in the camera image shown in Fig. \ref{fig:f2} (b).
\subsubsection{Feature Embedding}
\begin{figure}
    \centering
    \includegraphics[width=1\linewidth]{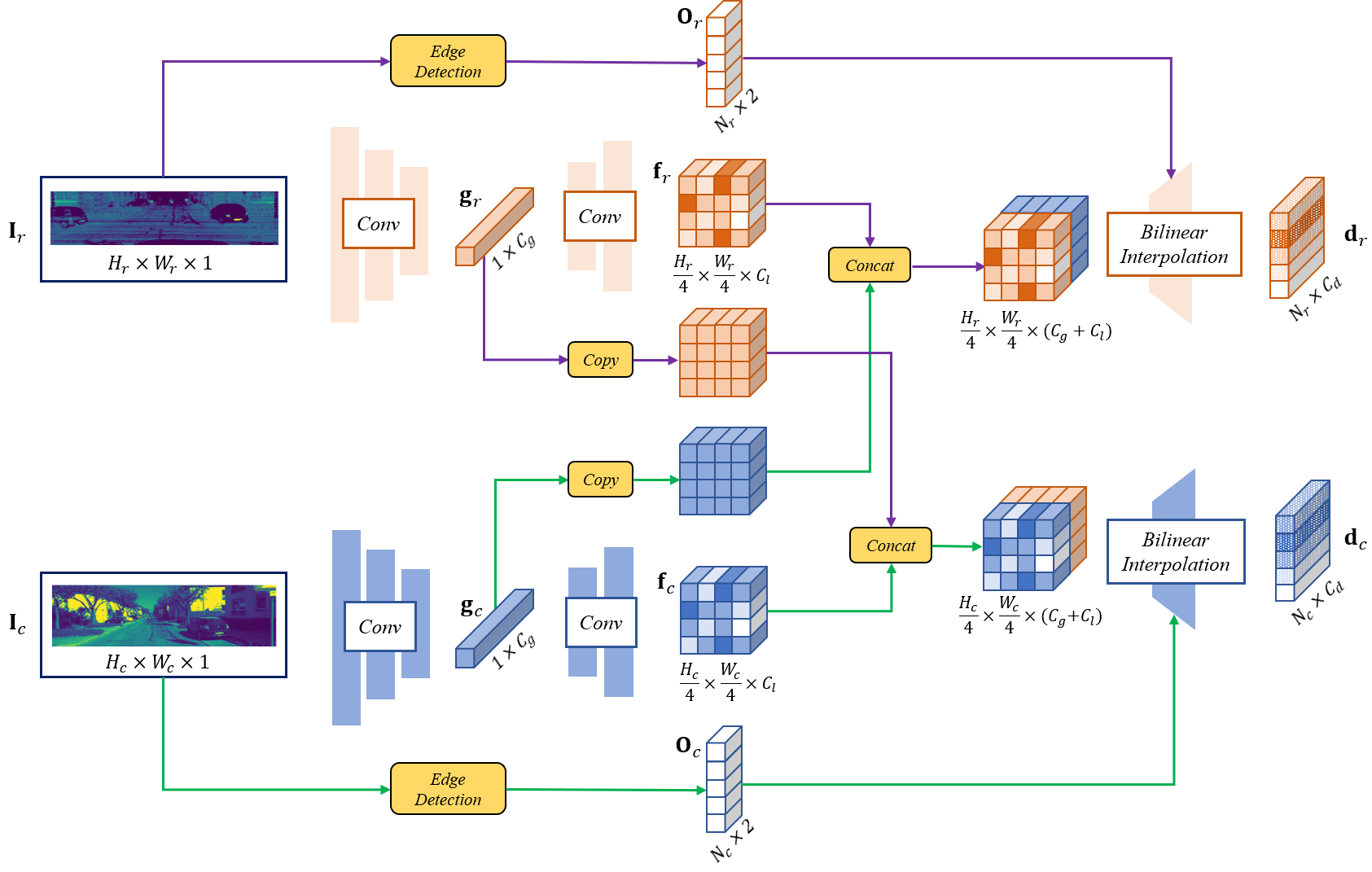}
    \caption{Neural network structure for feature extraction. The network takes reflectance map \(\textbf{I}_r\) and camera image \(\textbf{I}_c\) as inputs. Edge pixel extraction is performed on both inputs to obtain sequences of respective edge pixels. The output of this network architecture consists of features \(\textbf{d}_r\) corresponding to edge pixels in the reflectance map and features \(\textbf{d}_c\) corresponding to edge pixels in the camera image.}
    \label{fig:f3}
\end{figure}

Our feature extraction network, depicted in Fig. \ref{fig:f3}, comprises two symmetrical branches. Each branch receives inputs from the reflectance map \(\mathbf{I}_r \in \mathbb{R}^{H_r \times W_r \times 1}\), obtained through projection, and a single-channel image \(\mathbf{I}_c \in \mathbb{R}^{H_c \times W_c \times 1}\) capturing the red channel of the RGB image captured by the camera. The reason for using only the red channel of the camera image is that the light emitted by the LiDAR scanner is typically in the infrared spectrum, making the red light component of the camera image more closely resemble the reflectance situation in the reflectance map. The red channel of the camera image is illustrated in Fig. \ref{fig:f2} (a). Employing single-channel inputs not only enhances the similarity between the inputs of the two branches but also saves computational resources.\\
Each branch initially uses a ResNet encoder to extract global features \(\mathbf{g}_r \in \mathbb{R}^{C_g \times 1}\) and \(\mathbf{g}_c \in \mathbb{R}^{C_g \times 1}\) from the reflectance map and camera image, respectively, where \(C_g\) denotes the number of channels for global features. These global features encapsulate the global information of the entire image. \\Subsequently, ResNet decoders are used to process the global features, yielding local features \(\mathbf{f}_r \in \mathbb{R}^{\frac{H_r}{4} \times \frac{W_r}{4} \times C_l}\) and \(\mathbf{f}_c \in \mathbb{R}^{\frac{H_c}{4} \times \frac{W_c}{4} \times C_l}\) for the reflectance map and the camera image, respectively, where \(C_l\) represents the number of channels for local features. Although the two branches adopt identical feature extraction methods, differences in input data persist. To address this, we concatenate the obtained local features with the global features from the other branch, i.e., \(\mathbf{f}_r\) concatenated with \(\mathbf{g}_c\) and \(\mathbf{f}_c\) concatenated with \(\mathbf{g}_r\), facilitating information exchange between features and yielding fused features with a channel dimension of \(C_l + C_g\).\\
We employ bilinear interpolation to aggregate the fused features onto the edge pixels extracted in Section 3.2.1, represented as \(\mathbf{O}_r\) and \(\mathbf{O}_c \). This technique facilitates the extraction of features corresponding to edge pixels in both the reflectance map and the camera image, yielding \(\mathbf{d}_r \in \mathbb{R}^{N_r \times C_d}\) and \(\mathbf{d}_c \in \mathbb{R}^{N_c \times C_d}\), respectively.
\subsection{Matching Optimization Network}
\begin{figure}
    \centering
    \includegraphics[width=0.75\linewidth]{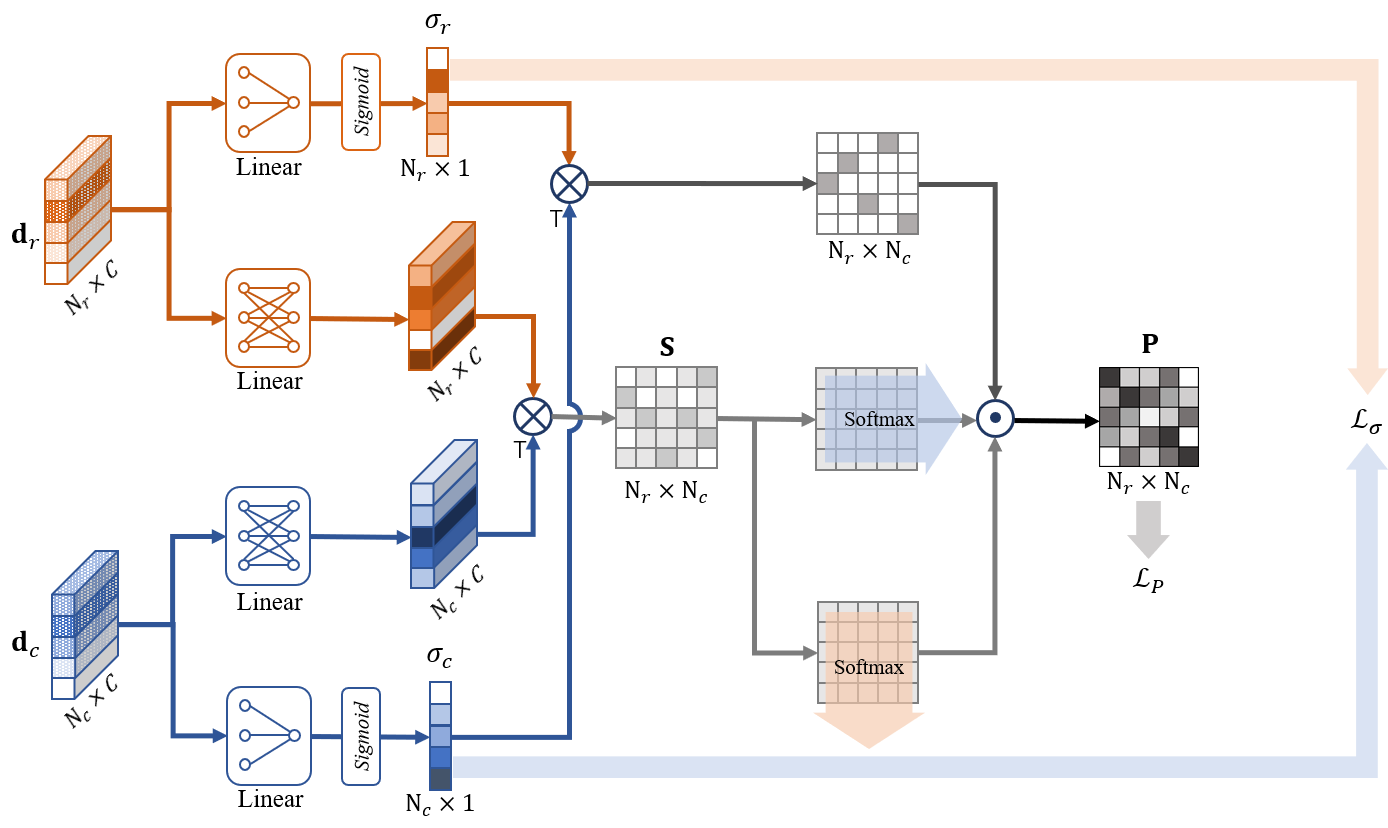}
    \caption{Structure of the Matching Optimization Network, illustrating the process from point features \( \mathbf{d}_r \) and \( \mathbf{d}_c \) to the correspondence decision via the partial assignment matrix \( \mathbf{P} \). Linear blocks represent linear neural network layers with varying output dimensions, while Softmax operations along matrix rows and columns are indicated by differently colored arrows.}
        \label{fig:f4}
\end{figure}

Upon acquiring features \( \mathbf{d}_r \) and \( \mathbf{d}_c \) for the edge pixels, the next step is to establish pixel correspondences based on the similarity of these features. Traditionally, cosine similarity matrices serve as a common tool for this purpose. SuperGlue\cite{sarlin2020superglue} addresses the assignment problem by utilizing optimal transport\cite{MAL-073} and the Sinkhorn algorithm\cite{Sinkhorn1967ConcerningNM}, significantly improving the point matching performance. To further enhance efficiency, LightGlue\cite{lindenberger2023lightglue} introduces a more efficient matching optimization layer to solve the assignment problem. Drawing inspiration from LightGlue, we propose a matching optimization layer to refine the precision of point matching. The structure of this optimization layer is depicted in Fig. \ref{fig:f4}.
To begin with, we subject the features to a learned linear transformation with matching input and output dimensions, followed by matrix multiplication to produce the pairwise score matrix \(\mathbf{S} \in \mathbb{R}^{N_r \times N_c}\) between the edge pixels of the reflectance map and the camera image. Mathematically, this operation is succinctly expressed as

\[
\mathbf{S} = \text{Linear}(\mathbf{d}_r)\times\text{Linear}(\mathbf{d}_c)^{\mathsf{T}}.\tag{6}
\]

Subsequently, we process the features \( \mathbf{d}_r \) and \( \mathbf{d}_c \) through separate linear layers, each with output dimensions of 1. Following this, we apply the Sigmoid function to compute the matching scores \( \sigma_r \) and \( \sigma_c \) for each edge pixel of the reflectance map and the camera image, respectively.

Upon obtaining the pairwise score matrix \( \mathbf{S} \), we employ the softmax function individually along the rows and columns. The resulting softmax values are then combined and multiplied by the matching scores \( \sigma_r \) and \( \sigma_c \) to derive the partial assignment matrix \( \mathbf{P} \).

We determine 2D-2D matching pairs by checking if the column coordinates with maximum values in the partial assignment matrix \( \mathbf{P} \) also have maximum values within their respective columns. Upon identifying the 2D-2D matches, we establish the correspondence between the points in the LiDAR point cloud and pixels in the camera image based on the pair mapping recorded during the projection process in Section 3.1.
\subsection{Loss Function}
In the training phase, the loss function is divided into two parts, \(\mathcal{L}_\sigma\) and \(\mathcal{L}_P\). First, the 3D key points are projected onto the camera plane based on the actual transformation matrix and camera intrinsic parameters. Subsequently, the input 3D point cloud is projected onto a 2D image plane, and a circular radius threshold \( \epsilon \) is set. If the Euclidean distance between the projected point of a 3D key point and a 2D key point is less than \( \epsilon \), they are considered valid 2D-3D matching pairs. 
We denote whether the 2D and 3D points are matching points using the matching scores \( \sigma_{r_{GT}} \) and \( \sigma_{c_{GT}} \), respectively. In our task, the overlapping region between the camera image and the reflectance map is smaller compared to typical image-matching tasks, therefore, we heighten the precision requirements for the matching scores \( \sigma_r \) and \( \sigma_c \). To address this, we employ a binary cross-entropy loss function, aiming to improve the accuracy of matching pairs. Consequently, we obtain \( \mathcal{L}_{\sigma1} \) and \( \mathcal{L}_{\sigma2} \) as follows
\[ \mathcal{L}_{\sigma1} = -\{\sigma_{r_{GT}} \cdot \log{\left(\sigma_r\right)} + \left(1-\sigma_{r_{GT}}\right) \cdot \log{\left(1-\sigma_r\right)}\},\tag{7} \]
\[ \mathcal{L}_{\sigma2} = -\{\sigma_{c_{GT}} \cdot \log{\left(\sigma_c\right)} + \left(1-\sigma_{c_{GT}}\right) \cdot \log{\left(1-\sigma_c\right)}\}.\tag{8} \]
The sum of these two losses yields \( \mathcal{L}_\sigma = \mathcal{L}_{\sigma1} + \mathcal{L}_{\sigma2} \).
For the partial assignment matrix \( \mathbf{P} \), the positions corresponding to the maximum values in both rows and columns are considered as predicted 2D-3D matching pairs. After obtaining the actual 2D-3D matching pairs, the values corresponding to the actual matching positions in the partial assignment matrix \( \mathbf{P} \) should be relatively large. Based on this principle, we calculate 
\[ \mathcal{L}_P = -\frac{1}{\mathcal{M}}\sum\limits_{(i,j) \in \mathcal{M}} \log \mathbf{P}_{ij} , \tag{9}\]
where \( \mathcal{M} \) represents the positions of actual matching pairs. The overall loss function is the sum of these two parts, \( \mathcal{L} = \mathcal{L}_\sigma + \mathcal{L}_P \).
\subsection{Transformation Matrix Estimation}
To ensure the applicability and robustness of the proposed method, we opted not to use an end-to-end pose regression network. Instead, we employ the EPnP algorithm combined with RANSAC\cite{10.1145/358669.358692} to estimate the transformation matrix directly. An additional point to note is that our matching optimization layer ensures precise alignment with 2D-2D matches between the edge pixels of the reflectance map and those of the camera image.  To undertake EPnP estimation effectively, a crucial step involves mapping the edge pixels in the reflectance map to their corresponding 3D points within the point cloud, facilitated by the projection matrix. This mapping relationship was elaborated upon in Section 3.1. Through this mapping process, we seamlessly transition the 2D-2D correspondence to a 3D-2D correspondence between the LiDAR and the camera, thereby enabling the EPnP algorithm to estimate the transformation matrix accurately.
\section{Experiments}
\subsection{Dataset}
We trained and tested the proposed network on the KITTI Odometry dataset\cite{6248074} by using the same-frame 2D/3D sensor data to generate image point cloud pairs. For the KITTI dataset, we followed previous methodologies by using sequences 0-8 for training and sequences 9-10 for testing. Artificial errors applied to the point clouds during training and testing include ±10m two-dimensional translations along the \textit{x} and \textit{y} axes as well as arbitrary rotations around the \textit{z}-axis. It is worth noting that, since the center position in the spherical coordinates of point clouds is typically fixed, translations along the \textit{x} and \textit{y} axes applied in other works do not accurately reflect real-world scenarios. To ensure fairness and realism, we initially applied arbitrary rotations around the \textit{z}-axis to the point cloud before generating reflectance maps. Subsequently, after generating the reflectance maps, we applied translations along the \textit{x} and \textit{y} axes to the point cloud. We downsample the camera image resolution to 160×512 and employ reflectance maps with a resolution of 64×1024 for both training and testing. 
\subsection{Baselines and Metrics}
We compared our method with the state-of-the-art approaches DeepI2P\cite{li2021deepi2p}, CorrI2P\cite{Ren_2023}, and VP2P-Match\cite{NEURIPS2023_a0a53fef}:\\
1) \textbf{DeepI2P.} Two methods were proposed in DeepI2P, namely \textbf{Grid Cls. + PnP} and \textbf{Frus. Cls. + Inv. Proj.}, respectively. \textbf{Grid Cls. + PnP} divides an input image into a 32×32 grid and trains a neural network to classify 3D points into specific 2D grid cells. Then it uses the EPnP algorithm combined with RANSAC to estimate the rigid transformation. \textbf{Frus.Cls. + Inv. Proj.} introduces frustum classification using inverse camera projection to determine the rigid transformation. It explores both 2D and 3D inverse camera projection methods, denoted as DeepI2P(2D) and DeepI2P(3D), respectively.\\
2) \textbf{CorrI2P.} Developed after DeepI2P, CorrI2P improves the accuracy of the image-to-point cloud registration. It employs a neural network with a feature exchange module and supervision of overlapping regions to learn dense correspondences between image and point cloud pairs. The EPnP algorithm with RANSAC is used to predict rigid transformations, enhancing registration performance.\\
3) \textbf{VP2P-Match.} VP2P-Match enhances pixel-point matching accuracy and speed by leveraging sparse convolution to increase the similarity between point cloud and CNN-extracted image features. It integrates a differentiable PnP solver into an end-to-end training schema, enabling learning of a structured cross-modality latent space with adaptive-weighted optimization for Image-to-Point Cloud registration.
\subsection{Registration Accuracy}
Similar to point-to-point (P2P) registration, we used the relative translational error (RTE) \(E_t\) and the relative rotational error (RRE) \(E_R\) to evaluate the registration errors:

\[E_R = \sum_{i=1}^{3} \left| \gamma(i) \right|,\tag{10}\]
\[E_t = \left\| \mathbf{t}_{gt} - \mathbf{t}_E \right\|,\tag{11}\]

where \(\gamma\) is the Euler angle of the matrix \(\mathbf{R}_{gt}^{-1}\mathbf{R}_E\), \(\mathbf{R}_{gt}\), and \(\mathbf{t}_{gt}\) are the transformation of the ground truth, and \(\mathbf{R}_E\) and \(\mathbf{t}_E\) are the estimated transformation.
Similar to VP2P-Match, we calculated the mean and standard deviation of RRE (Relative Rotation Error) and RTE (Relative Translation Error) for all test data. Additionally, we evaluated the registration success rate (Acc.) based on the criteria of RRE < 5° and RTE < 2m.
\begin{table}[H]
\centering
\caption{Registration accuracy on the KITTI dataset. Lower is better for RTE and RRE, higher is better for Acc.}
\begin{tabular}{l|c|c|c}
\hline
Method          & RTE(m) ↓    & RRE(°) ↓    & Acc.↑ \\ \hline
Grid Cls. + EPnP \cite{li2021deepi2p}& 3.64 ± 3.46& 19.19 ± 28.96& 11.22 \\
DeepI2P (3D) \cite{li2021deepi2p}& 4.06 ± 3.54& 24.73 ± 31.69& 3.77  \\
DeepI2P (2D) \cite{li2021deepi2p}& 3.59 ± 3.21& 11.66 ± 18.16& 25.95 \\
CorrI2P \cite{Ren_2023}& 3.78 ± 65.16& 5.89 ± 20.34& 72.42 \\
VP2P-Match \cite{NEURIPS2023_a0a53fef}& 0.75 ± 1.13& 3.29 ± 7.99& 83.04 \\
Ours            & \textbf{0.54 ± 0.85}& \textbf{2.97 ± 3.05}& \textbf{85.74}\\ \hline
\end{tabular}
\end{table}

\subsection{Complexity Comparision}
\begin{wraptable}{r} {5.8cm}
\vspace{-20pt}
\centering
\caption{Complexity Comparison of different methods.}
\label{t2}

\begin{tabular}{c|c}
\hline
Method           & Inference Time(s)\\ \hline
Grid Cls. + EPnP & 0.04              \\
DeepI2P(2D)      & 16.58             \\
DeepI2P(3D)      & 9.38              \\
CorrI2P          & 2.97              \\
VP2P-Match       & 0.19              \\
Ours             & \textbf{0.015}\\ \hline
\end{tabular}
\end{wraptable}
We compared the complexity of the proposed methods with others on a platform equipped with an NVIDIA GeForce RTX 3090 GPU and an Intel(R) Core(TM) i9-10900X CPU @ 3.70GHz, as shown in Table \ref{t2}. In the test, we measured the average time per scene during network testing by dividing the total testing time by the number of scenes in the test set. Due to variations in testing environments and procedures, we compared our results with the shortest times recorded for each method in the existing literature.

\subsection{Ablation Study}
\textbf{Ablation Study on Reflectance Map and Depth Map. }During projection, we can project the reflectance information or depth information of point clouds to obtain different forms of projection maps. To demonstrate the superior performance of the reflectance maps, compared to depth maps that rely only on spatial information, we conducted ablation experiments. The results, presented in Table \ref{t3}, show that using depth maps leads to limited performance in scenarios where spatial information within the camera's field of view is insufficient. In contrast, our approach utilizing reflectance maps consistently performs well across various scenarios. \\
\textbf{Ablation Study on the Reflenctance Map with Different Resolutions.} We also conducted ablation experiments to examine the impact of different resolution projection images on performance, as shown in Table 3. The results indicate that selecting an appropriate size for the projection images can enhance the network's performance.
\begin{table}[H]
\centering
\caption{Ablation study at different conditions: the use of depth maps and the different sizes of reflectance maps.}
\label{t3}
\begin{tabular}{l|c|c|c|c}
\hline
Input Categories& Size     & RTE(m) ↓    & RRE(°) ↓    & Acc.↑ \\ \hline
Depth Map                        & 64×1024  & 0.81 ± 2.20 & 3.43 ± 8.34 & 81.52 \\ \hline
\multirow{3}{*}{Reflectance Map} & 32×512   & 0.67 ± 1.21 & 3.36 ± 4.57 & 81.59 \\
                                 & 128×2048 &             0.72 ± 0.71&             3.37 ± 3.35&       80.96\\
                                 & 64×1024  & \textbf{0.54 ± 0.85}& \textbf{2.97 ± 3.05}& \textbf{85.74}\\ \hline
\end{tabular}
\end{table}
\section{Conclusions}
In this work, we proposed EEPNet, an efficient real-time registration method between LiDAR point clouds and camera images by using  reflectance projection. Our approach leverages reflectance maps and edge information to enhance the registration process, achieving superior performance in terms of both accuracy and efficiency. However, the algorithm's performance may degrade in scenarios where the extracted 2D and 3D key points are limited. Future work could focus on optimizing the extraction of key points to enhance the algorithm's robustness and applicability in such challenging conditions.

\medskip
\bibliographystyle{ieee}
\bibliography{references}

\begin{thebibliography}{10}\itemsep=-1pt

\bibitem{9157624}
X.~Bai, Z.~Luo, L.~Zhou, H.~Fu, L.~Quan, and C.-L. Tai.
\newblock D3feat: Joint learning of dense detection and description of 3d local features.
\newblock In {\em 2020 IEEE/CVF Conference on Computer Vision and Pattern Recognition (CVPR)}, pages 6358--6366, 2020.

\bibitem{BAY2008346}
H.~Bay, A.~Ess, T.~Tuytelaars, and L.~{Van Gool}.
\newblock Speeded-up robust features (surf).
\newblock {\em Computer Vision and Image Understanding}, 110(3):346--359, 2008.
\newblock Similarity Matching in Computer Vision and Multimedia.

\bibitem{121791}
P.~Besl and N.~D. McKay.
\newblock A method for registration of 3-d shapes.
\newblock {\em IEEE Transactions on Pattern Analysis and Machine Intelligence}, 14(2):239--256, 1992.

\bibitem{1249285}
P.~Biber and W.~Strasser.
\newblock The normal distributions transform: a new approach to laser scan matching.
\newblock In {\em Proceedings 2003 IEEE/RSJ International Conference on Intelligent Robots and Systems (IROS 2003) (Cat. No.03CH37453)}, volume~3, pages 2743--2748 vol.3, 2003.

\bibitem{detone2018superpoint}
D.~DeTone, T.~Malisiewicz, and A.~Rabinovich.
\newblock Superpoint: Self-supervised interest point detection and description, 2018.

\bibitem{8794415}
M.~Feng, S.~Hu, M.~H. Ang, and G.~H. Lee.
\newblock 2d3d-matchnet: Learning to match keypoints across 2d image and 3d point cloud.
\newblock In {\em 2019 International Conference on Robotics and Automation (ICRA)}, pages 4790--4796, 2019.

\bibitem{10.1145/358669.358692}
M.~A. Fischler and R.~C. Bolles.
\newblock Random sample consensus: a paradigm for model fitting with applications to image analysis and automated cartography.
\newblock {\em Commun. ACM}, 24(6):381–395, jun 1981.

\bibitem{5563693}
W.~Gao, X.~Zhang, L.~Yang, and H.~Liu.
\newblock An improved sobel edge detection.
\newblock In {\em 2010 3rd International Conference on Computer Science and Information Technology}, volume~5, pages 67--71, 2010.

\bibitem{6248074}
A.~Geiger, P.~Lenz, and R.~Urtasun.
\newblock Are we ready for autonomous driving? the kitti vision benchmark suite.
\newblock In {\em 2012 IEEE Conference on Computer Vision and Pattern Recognition}, pages 3354--3361, 2012.

\bibitem{Grigorescu_2019}
S.~Grigorescu, B.~Trasnea, T.~Cocias, and G.~Macesanu.
\newblock A survey of deep learning techniques for autonomous driving.
\newblock {\em Journal of Field Robotics}, 37(3):362–386, Nov. 2019.

\bibitem{10160910}
Y.~Hu, H.~Ma, L.~Jie, and H.~Zhang.
\newblock Dedgenet: Extrinsic calibration of camera and lidar with depth-discontinuous edges.
\newblock In {\em 2023 IEEE International Conference on Robotics and Automation (ICRA)}, pages 11439--11445, 2023.

\bibitem{8593693}
G.~Iyer, R.~K. Ram, J.~K. Murthy, and K.~M. Krishna.
\newblock Calibnet: Geometrically supervised extrinsic calibration using 3d spatial transformer networks.
\newblock In {\em 2018 IEEE/RSJ International Conference on Intelligent Robots and Systems (IROS)}, pages 1110--1117, 2018.

\bibitem{5970711}
X.~Jia, Z.~Hu, and H.~Guan.
\newblock A new multi-sensor platform for adaptive driving assistance system (adas).
\newblock In {\em 2011 9th World Congress on Intelligent Control and Automation}, pages 1224--1230, 2011.

\bibitem{10099031}
J.~Jiao, F.~Chen, H.~Wei, J.~Wu, and M.~Liu.
\newblock Lce-calib: Automatic lidar-frame/event camera extrinsic calibration with a globally optimal solution.
\newblock {\em IEEE/ASME Transactions on Mechatronics}, 28(5):2988--2999, 2023.

\bibitem{Lepetit2009}
V.~Lepetit, F.~Moreno-Noguer, and P.~Fua.
\newblock Epnp: An accurate o(n) solution to the pnp problem.
\newblock {\em International Journal of Computer Vision}, 81(2):155--166, Feb 2009.

\bibitem{10321736}
H.~Li, C.~Sima, J.~Dai, W.~Wang, L.~Lu, H.~Wang, J.~Zeng, Z.~Li, J.~Yang, H.~Deng, H.~Tian, E.~Xie, J.~Xie, L.~Chen, T.~Li, Y.~Li, Y.~Gao, X.~Jia, S.~Liu, J.~Shi, D.~Lin, and Y.~Qiao.
\newblock Delving into the devils of bird’s-eye-view perception: A review, evaluation and recipe.
\newblock {\em IEEE Transactions on Pattern Analysis and Machine Intelligence}, 46(4):2151--2170, 2024.

\bibitem{li2021deepi2p}
J.~Li and G.~H. Lee.
\newblock Deepi2p: Image-to-point cloud registration via deep classification, 2021.

\bibitem{lindenberger2023lightglue}
P.~Lindenberger, P.-E. Sarlin, and M.~Pollefeys.
\newblock Lightglue: Local feature matching at light speed, 2023.

\bibitem{790410}
D.~Lowe.
\newblock Object recognition from local scale-invariant features.
\newblock In {\em Proceedings of the Seventh IEEE International Conference on Computer Vision}, volume~2, pages 1150--1157 vol.2, 1999.

\bibitem{lv2021lccnet}
X.~Lv, B.~Wang, D.~Ye, and S.~Wang.
\newblock Lccnet: Lidar and camera self-calibration using cost volume network, 2021.

\bibitem{8633982}
X.~Mao, W.~Li, C.~Lei, J.~Jin, F.~Duan, and S.~Chen.
\newblock A brain–robot interaction system by fusing human and machine intelligence.
\newblock {\em IEEE Transactions on Neural Systems and Rehabilitation Engineering}, 27(3):533--542, 2019.

\bibitem{pautrat2023gluestick}
R.~Pautrat, I.~Suárez, Y.~Yu, M.~Pollefeys, and V.~Larsson.
\newblock Gluestick: Robust image matching by sticking points and lines together, 2023.

\bibitem{pavlov2017aaicp}
A.~L. Pavlov, G.~V. Ovchinnikov, D.~Y. Derbyshev, D.~Tsetserukou, and I.~V. Oseledets.
\newblock Aa-icp: Iterative closest point with anderson acceleration, 2017.

\bibitem{MAL-073}
G.~Peyré and M.~Cuturi.
\newblock Computational optimal transport: With applications to data science.
\newblock {\em Foundations and Trends® in Machine Learning}, 11(5-6):355--607, 2019.

\bibitem{pham2019lcd}
Q.-H. Pham, M.~A. Uy, B.-S. Hua, D.~T. Nguyen, G.~Roig, and S.-K. Yeung.
\newblock Lcd: Learned cross-domain descriptors for 2d-3d matching, 2019.

\bibitem{Ren_2023}
S.~Ren, Y.~Zeng, J.~Hou, and X.~Chen.
\newblock Corri2p: Deep image-to-point cloud registration via dense correspondence.
\newblock {\em IEEE Transactions on Circuits and Systems for Video Technology}, 33(3):1198–1208, Mar. 2023.

\bibitem{6126544}
E.~Rublee, V.~Rabaud, K.~Konolige, and G.~Bradski.
\newblock Orb: An efficient alternative to sift or surf.
\newblock In {\em 2011 International Conference on Computer Vision}, pages 2564--2571, 2011.

\bibitem{sarlin2020superglue}
P.-E. Sarlin, D.~DeTone, T.~Malisiewicz, and A.~Rabinovich.
\newblock Superglue: Learning feature matching with graph neural networks, 2020.

\bibitem{7995968}
N.~Schneider, F.~Piewak, C.~Stiller, and U.~Franke.
\newblock Regnet: Multimodal sensor registration using deep neural networks.
\newblock In {\em 2017 IEEE Intelligent Vehicles Symposium (IV)}, pages 1803--1810, 2017.

\bibitem{Sinkhorn1967ConcerningNM}
R.~Sinkhorn and P.~Knopp.
\newblock Concerning nonnegative matrices and doubly stochastic matrices.
\newblock {\em Pacific Journal of Mathematics}, 21:343--348, 1967.

\bibitem{9802778}
Y.~Sun, J.~Li, Y.~Wang, X.~Xu, X.~Yang, and Z.~Sun.
\newblock Atop: An attention-to-optimization approach for automatic lidar-camera calibration via cross-modal object matching.
\newblock {\em IEEE Transactions on Intelligent Vehicles}, 8(1):696--708, 2023.

\bibitem{9010002}
H.~Thomas, C.~R. Qi, J.-E. Deschaud, B.~Marcotegui, F.~Goulette, and L.~Guibas.
\newblock Kpconv: Flexible and deformable convolution for point clouds.
\newblock In {\em 2019 IEEE/CVF International Conference on Computer Vision (ICCV)}, pages 6410--6419, 2019.

\bibitem{10.1007/3-540-44480-7_21}
B.~Triggs, P.~F. McLauchlan, R.~I. Hartley, and A.~W. Fitzgibbon.
\newblock Bundle adjustment --- a modern synthesis.
\newblock In B.~Triggs, A.~Zisserman, and R.~Szeliski, editors, {\em Vision Algorithms: Theory and Practice}, pages 298--372, Berlin, Heidelberg, 2000. Springer Berlin Heidelberg.

\bibitem{Wang_2021_ICCV}
B.~Wang, C.~Chen, Z.~Cui, J.~Qin, C.~X. Lu, Z.~Yu, P.~Zhao, Z.~Dong, F.~Zhu, N.~Trigoni, and A.~Markham.
\newblock P2-net: Joint description and detection of localfeaturesfor pixel and point matching.
\newblock In {\em Proceedings of the IEEE/CVF International Conference on Computer Vision (ICCV)}, pages 16004--16013, October 2021.

\bibitem{wang2023endtoend}
G.~Wang, Y.~Zheng, Y.~Guo, Z.~Liu, Y.~Zhu, W.~Burgard, and H.~Wang.
\newblock End-to-end 2d-3d registration between image and lidar point cloud for vehicle localization, 2023.

\bibitem{electronics10111224}
T.~Wu, H.~Fu, B.~Liu, H.~Xue, R.~Ren, and Z.~Tu.
\newblock Detailed analysis on generating the range image for lidar point cloud processing.
\newblock {\em Electronics}, 10(11), 2021.

\bibitem{6751291}
J.~Yang, H.~Li, and Y.~Jia.
\newblock Go-icp: Solving 3d registration efficiently and globally optimally.
\newblock In {\em 2013 IEEE International Conference on Computer Vision}, pages 1457--1464, 2013.

\bibitem{9623545}
C.~Ye, H.~Pan, and H.~Gao.
\newblock Keypoint-based lidar-camera online calibration with robust geometric network.
\newblock {\em IEEE Transactions on Instrumentation and Measurement}, 71:1--11, 2022.

\bibitem{9206138}
K.~Yuan, Z.~Guo, and Z.~J. Wang.
\newblock Rggnet: Tolerance aware lidar-camera online calibration with geometric deep learning and generative model.
\newblock {\em IEEE Robotics and Automation Letters}, 5(4):6956--6963, 2020.

\bibitem{zhang2023lidargenerated}
H.~Zhang, X.~Yu, S.~Ha, and T.~Westerlund.
\newblock Lidar-generated images derived keypoints assisted point cloud registration scheme in odometry estimation, 2023.

\bibitem{NEURIPS2023_a0a53fef}
J.~Zhou, B.~Ma, W.~Zhang, Y.~Fang, Y.-S. Liu, and Z.~Han.
\newblock Differentiable registration of images and lidar point clouds with voxelpoint-to-pixel matching.
\newblock In A.~Oh, T.~Naumann, A.~Globerson, K.~Saenko, M.~Hardt, and S.~Levine, editors, {\em Advances in Neural Information Processing Systems}, volume~36, pages 51166--51177. Curran Associates, Inc., 2023.

\bibitem{8593660}
L.~Zhou, Z.~Li, and M.~Kaess.
\newblock Automatic extrinsic calibration of a camera and a 3d lidar using line and plane correspondences.
\newblock In {\em 2018 IEEE/RSJ International Conference on Intelligent Robots and Systems (IROS)}, pages 5562--5569, 2018.

\bibitem{zhou2022geometry}
Q.~Zhou, S.~Agostinho, A.~Osep, and L.~Leal-Taixé.
\newblock Is geometry enough for matching in visual localization?, 2022.

\end{thebibliography}

\newpage
\appendix

\section{Appendix / supplemental material}
\subsection{Network Details}
\begin{figure}[H]
    \centering
    \includegraphics[width=1\linewidth]{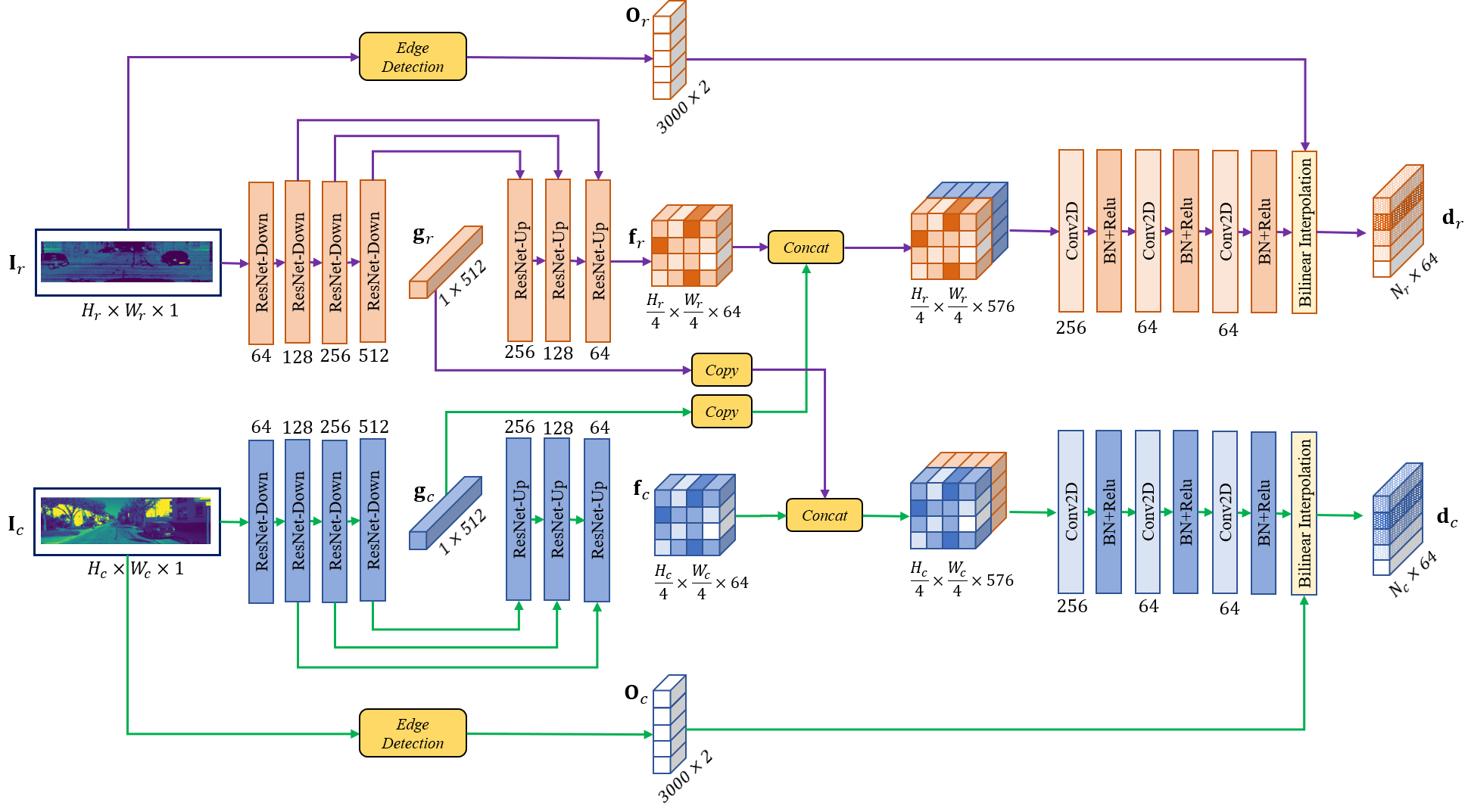}
    \caption{Detailed network structure of the feature extraction module in which the dimensions of the features are also indicated. Conv2D represents a 2D convolutional layer, BN denotes a batch normalization layer, and ReLU indicates the ReLU activation function. }
    \label{fig:sf1}
\end{figure}

In the feature matching network, the global feature dimension \(C_g\) is 512, and the local feature dimension \(C_l\) is 64. The final dimension of the edge pixels' feature \(C_d\) is 64, with the number of edge pixels on the reflectance map \(N_r\) set to 3000, and the number of edge pixels in the camera image \(N_c\) also set to 3000.\\
Fig. \ref{fig:sf1} illustrates the detailed architecture of the feature extraction network, which is identical for both the reflectance map \(\mathbf{I}_r\) and the camera image \(\mathbf{I}_c\). The ResNet34 used for image feature extraction comprises four layers of ResNet encoders, with each layer's output dimensions shown in the figure. After the encoding stage, the global features \(\mathbf{g}_r\) and \(\mathbf{g}_c\) have an output dimension of 512. These features are then upsampled using a ResNet34 decoder. The upsampling stage consists of three layers of decoder modules, which ultimately produce the features \(\mathbf{f}_r\) and \(\mathbf{f}_c\) with a scale of 1/4 of the original input and a dimension of 64.
Following the feature concatenation, a 2D convolution is applied to the concatenated features. This operation reduces the concatenated feature dimension from 576 to 64. After this, bilinear interpolation is performed, resulting in the aggregated features \(\mathbf{d}_r\) and \(\mathbf{d}_c\) on the edge pixels.
\subsection{Training Details}
We used the Adam optimizer with a learning rate of 0.001. The batch size was set to 16, and the number of epochs was set to 30. The threshold radius to determine whether a pair is a match is set to \(\epsilon = 2\). For the reprojection error of the EPnP algorithm, which we denote as \(\epsilon_e\), we discuss its impact on accuracy and efficiency in Section A.4.2 Ultimately, we set \(\epsilon_e = 6\).\\
We did not crop the first 50 rows of pixels from the images, as done in other studies, which justified this by claiming that these portions mostly contained useless sky information. We believe that such cropping is not suitable for better simulating real-world conditions.
We applied a wavelet filter to the reflectivity maps, whereas the camera images did not undergo the wavelet filter to better simulate real-world conditions. Specifically, after performing the wavelet transform on the reflectance maps, we set a wavelet filter threshold of 50, where high-frequency coefficients exceeding this threshold were smoothed. For the Sobel operator, the threshold for the camera image is set to 80, and for the reflectance map, it is set to 50. 
\subsection{Visualization of registration Results}
\begin{figure}[H]
    \centering
    \includegraphics[width=1\linewidth]{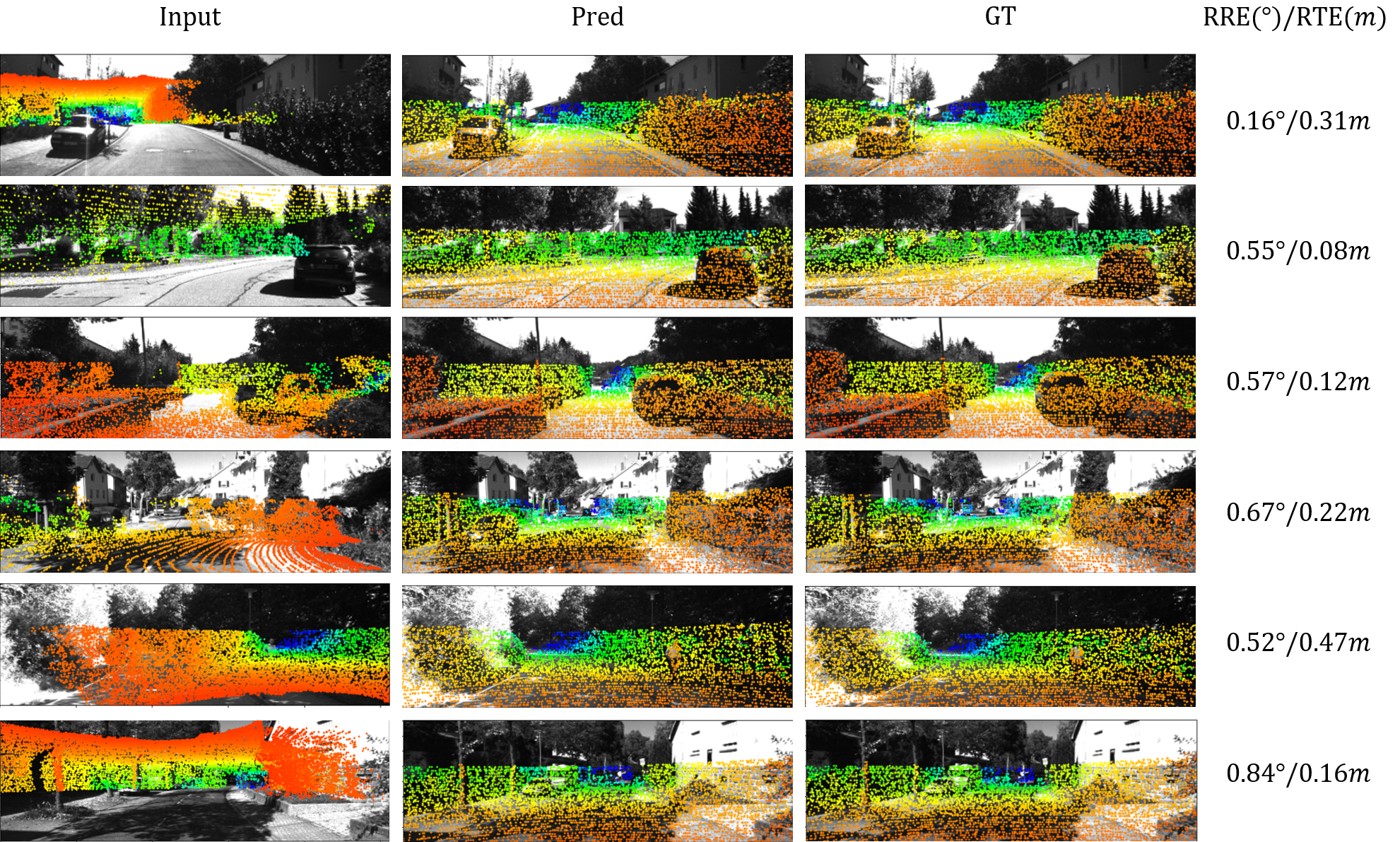}
    \caption{Visualization of registration results. The images show the effects of projecting the point cloud onto the image plane using different transformation matrices. 'Input' represents the unregistered network input, 'Pred' represents the registration results predicted by the network, and 'GT' represents the registration results obtained using Ground Truth parameters. The RRE and RTE values for this set of experimental data are shown on the right side of the image.}
    \label{fig:sf2}
\end{figure}

To provide a clearer demonstration, we uniformly downsampled the point cloud to 40,000 points. Fig. \ref{fig:sf2} shows the visualization results of projecting the point cloud onto the image plane using different extrinsic parameters. The background is the grayscale image of the same frame from the camera, and the color of the point cloud represents the depth in the spherical coordinate system.

\subsection{Additional Experiments and Analysis }

\subsubsection{Analysis of Specific Scenarios }
\begin{figure}[H]
    \centering
    \includegraphics[width=1\linewidth]{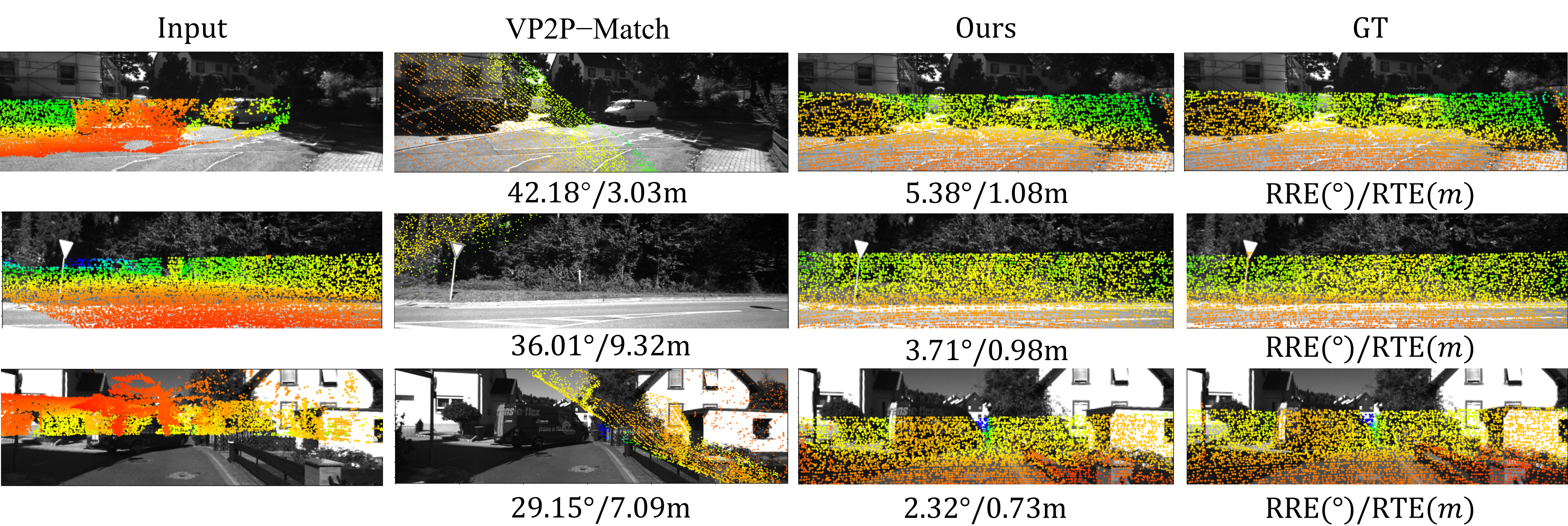}
    \caption{Visualization analysis of specific scenarios where the spatial information of the point cloud within the camera's field of view is not sufficiently rich. 'VP2P-Match' represents the registration results obtained by the VP2P-Match method in this scenario, while 'Ours' represents the results obtained by our method. Similarly, results for RRE and RTE are provided below each image. }
    \label{fig:sf3}
\end{figure}

We mentioned that our method leverages reflectance information for registration, which allows it to perform well even when the spatial features of the point cloud within the camera's field of view are not prominent. To validate this claim, we conducted a detailed analysis of experimental results that meet these conditions.\\
We consider results with RRE > 10° or RTE > 5m as extreme registration failures and have calculated the proportion of such instances, referred to as the Bad Rate. The results are summarized in Table \ref{st1}. By comparing the Bad Rate with the VP2P-Match method, it is evident that our method demonstrates superior registration stability across various scenarios.
\begin{table}[H]
\centering
\caption{Comparison of Bad Rate Results.}
\label{st1}
\begin{tabular}{c|c}
\hline
Method     & Bad Rate(\%) \\ \hline
VP2P-Match & 4.94         \\
Ours       & 0.47         \\ \hline
\end{tabular}
\end{table}
We also conducted separate visual analyses for cases where the spatial information of the point cloud within the camera's field of view was not prominent. The visualization results are shown in Fig. \ref{fig:sf3}. Through this analysis, we aim to gain insights into the performance of our method under conditions where the spatial information of the point cloud within the camera's field of view is not sufficiently rich.

\subsubsection{The Analysis of Transformation Matrix Estimation}
To analyze the impact of the reprojection error threshold \(\epsilon_e\) in the EPnP algorithm with RANSAC, and to further investigate the effects of using reflectance images versus projection images as inputs under different conditions, we conducted ablation experiments. The results of these experiments are shown in Table\ref{st2}.
In the table, we report the RRE, RTE, and Acc values of the registration results under different conditions, reflecting the accuracy of the registration. "Total Time" indicates the time taken to process all pose inferences for the test set, while "Inference Time" represents the average time taken for pose inference in a single scene. \\From Table \ref{st2}, we can see that using depth maps consistently performs worse than using reflectance maps. The algorithm achieves a more balanced performance when \(\epsilon_e = 6\).
\begin{table}[H]
\centering
\caption{Comparison of the impact of \(\epsilon_e\) on registration results under different input image conditions.}
\label{st2}
\begin{tabular}{l|c|c|c|c|c|c}
\hline
Method                           & \(\epsilon_e\)& RTE(m)    & RRE(°)    & Acc.  & Total Time(s) & Inference Time(s)\\ \hline
\multirow{4}{*}{Reflectance Map} & 8.0            & 0.55±0.59 & 3.01±3.06 & 85.67 & 76.021        & 0.0136            \\
                                 & 6.0            & 0.54±0.85 & 2.97±3.05 & 85.74 & 84.802        & 0.0152            \\
                                 & 4.0            & 0.54±0.78 & 2.99±3.04 & 85.92 & 97.199        & 0.0174            \\
                                 & 2.0            & 0.57±0.45 & 3.08±1.97 & 83.75 & 98.931        & 0.0177            \\ \hline
\multirow{4}{*}{Depth Map}       & 8.0            & 0.85±2.55 & 3.45±7.86 & 80.93 & 81.698        & 0.0146            \\
                                 & 6.0            & 0.81±2.19 & 3.43±8.34 & 81.51 & 90.454        & 0.0162            \\
                                 & 4.0            & 0.77±1.66 & 3.33±6.00 & 80.48 & 98.842        & 0.0177            \\
                                 & 2.0            & 0.80±0.66 & 3.36±2.27 & 78.03 & 98.434        & 0.0176            \\ \hline
\end{tabular}
\end{table}
\subsubsection{Supplementary Analysis of Depth Maps }
In the experimental section, we conducted ablation studies using depth maps generated from projected depth information. To provide a more intuitive representation of the details in the depth maps we utilized, we supplement the visual results of the depth maps in Fig. \ref{fig:sf4}:
\begin{figure}[H]
    \centering
    \includegraphics[width=1\linewidth]{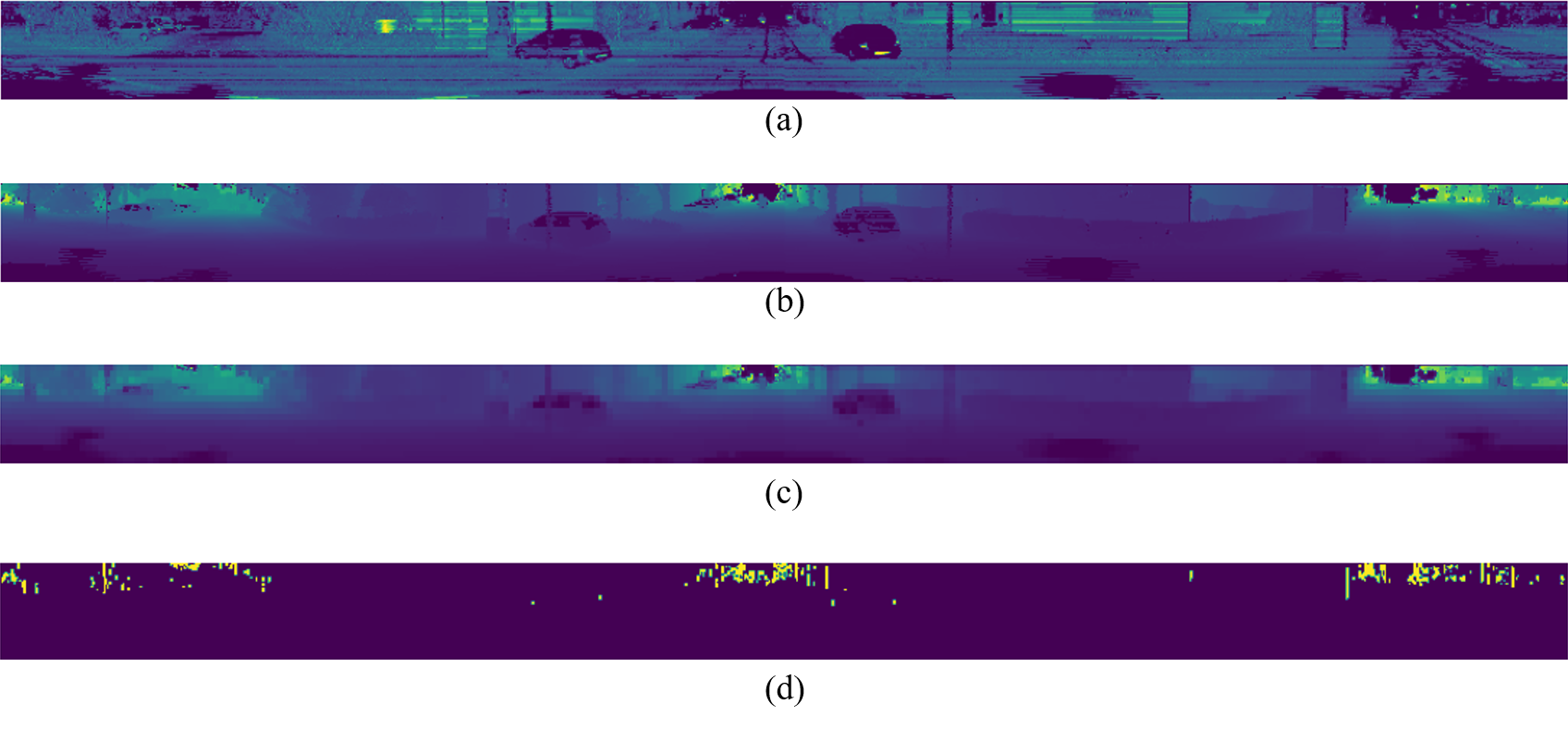}
    \caption{Visualization of Reflectance Map and Depth Map Processing. (a) Reflectance map obtained from point cloud projection. (b) Depth map obtained from point cloud projection. (c) Post-processing of the projection map with wavelet filter, derived from (b).  (d) Visualization of edge pixels extracted from (c). }
    \label{fig:sf4}
\end{figure}
In Figure \ref{fig:sf4}, panel (b) depicts the projection result of the depth map, which, compared to the reflectance map in Figure \ref{fig:sf4}(a), contains relatively less information. After post-processing with a wavelet filter, the depth map yields the image shown in panel (c) of Figure \ref{fig:sf4}. Subsequently, edge detection is applied to the processed image to obtain panel (d) in Figure \ref{fig:sf4}. It can be observed that the depth map extracts fewer edge pixels compared to the reflectance map.


\end{document}